%% file: main.tex
\RequirePackage{fix-cm}
\documentclass{article}
\PassOptionsToPackage{numbers,sort&compress}{natbib}
\usepackage[dblblindworkshop,final]{neurips_2026}
\usepackage[utf8]{inputenc}
\usepackage[T1]{fontenc}
\usepackage{courier}
\usepackage{amsmath,amsfonts}
\usepackage{graphicx,booktabs}
\usepackage{microtype}
\usepackage{xcolor}
\usepackage{tikz}
\usetikzlibrary{positioning,arrows.meta,calc}
\usepackage[hidelinks]{hyperref}
\usepackage{xurl}
\workshoptitle{AI for Drug Discovery: Bridging the Translation Gap}
\title{GeneSpeak-FP: Target and Compound Retrieval\\
from Observed Cell-Level Perturbation Signatures}
\author{%
  \begin{tabular}{@{}ccc@{}}
  Kseniia Vaniushkina & Jeongmin Lim & Jinyong Park \\
  {\normalfont\small\ttfamily seenear@toasterz.team} & {\normalfont\small\ttfamily ever\_since@jbnu.ac.kr} & {\normalfont\small\ttfamily jinyong5323@jbnu.ac.kr}
  \end{tabular}\\[0.4em]
  Correspondence: Kseniia Vaniushkina, \texttt{seenear@toasterz.team}
}
\hypersetup{pdftitle={GeneSpeak-FP: Target and Compound Retrieval from Observed Cell-Level Perturbation Signatures},pdfauthor={Kseniia Vaniushkina; Jeongmin Lim; Jinyong Park},pdfsubject={GeneSpeak-FP manuscript}}
\begin{document}
\maketitle
\input{paper_body}
\begin{ack}
This research received no specific funding. The authors declare no competing interests.
\end{ack}
\clearpage
\input{references}

\clearpage
\appendix
\input{appendices}
\end{document}

%% file: paper_body.tex
\begin{abstract}
Single-cell perturbation profiles may help identify known compounds across cellular contexts. We present GeneSpeak-FP, a gene-token Transformer that jointly ranks recorded compounds and their target annotations from a treated cell relative to a cell-line-specific DMSO reference, together with a coarse organ token. On Tahoe-100M, we split 11,673 drug--cell-line groups 90/10 within each compound. No complete group is shared across partitions, although compound and cell-line identities may recur. In a 38,400-query Monte Carlo pass, the correct compound appears first for 12.9\% of cells and within the top 10 for 34.3\%, among 379 fixed candidates. The model retrieves 40.8\% of annotated targets within the top 10 and 54.4\% within the top 20, from a vocabulary of 278 genes. Top-10 compound retrieval and target recall are approximately 13 and 11 times their expected values under uniformly random rankings; top-1 compound retrieval is nearly 49 times chance. Top-10 compound retrieval also exceeds the strongest evaluated post-hoc bag-of-genes control by more than nine times. A separate stochastic pass from the same validation pool gives nearly identical aggregate results. These findings support retrospective candidate-space narrowing in held-out drug--cell-line contexts. Because compound identities may recur and target labels are non-exhaustive metadata, they do not establish unseen-compound discovery, causal target identification, or clinical utility.
\end{abstract}

\section{Introduction}
Perturbational transcriptomics connects chemical interventions to downstream cell states. The Connectivity Map and L1000 established expression-signature search at scale \citep{ref4,ref14}, while single-cell screens expose response heterogeneity hidden by bulk averages \citep{ref11,ref13}. Tahoe-100M extends this setting to more than 100 million profiles across a broad cancer cell-line panel \citep{ref16}. These resources create an inverse question relevant to screen triage: given an observed response, which entries in a known compound library are most compatible with it?

The translational value of such retrieval depends on what the benchmark actually tests. A within-library split can reveal signal useful for organizing existing screens, yet it may also reward recognition of compounds already seen in other cell lines, technical covariates, or memorized metadata. Strong retrospective ranking is therefore evidence for candidate-space narrowing, not evidence that a model discovers a new molecule, identifies a causal target, or predicts efficacy. We make that evidence boundary part of the method report rather than treating it as a post-hoc disclaimer.

Prior work aligned bulk transcriptional profiles with molecular structures for cross-modal retrieval \citep{ref2}, derived perturbation barcodes for compound and target search \citep{ref1}, and built large-scale inverse signature search over LINCS \citep{ref8}. Forward single-cell models instead predict responses to specified interventions \citep{ref3,ref7,ref12}. Recent work has studied multimodal perturbation representations in single-cell data \citep{ref6,ref10}. Most directly for evaluation, Marrakchi et al.\ \citep{ref9} show that overlapping single-cell perturbation classes can make per-cell classification pessimistic even when population-level identification is strong.

GeneSpeak-FP contributes (i) a joint, cell-level retrieval model for recorded compounds and annotated targets in a fixed Tahoe-100M bank; (ii) a transparent audit of the split, candidate geometry, chance expectations, and label semantics; and (iii) an empirical feasibility result paired with an explicit set of tests needed to move from retrospective retrieval toward reliable drug-discovery use. We do not claim the first transcriptome-to-compound retrieval task or isolate the benefit of each architectural component.

\section{Closed-library retrieval}
\paragraph{Task and data.}
For treated cell $i$, compound $d$, cell line $c$, and gene index $g$, the query is an unpaired perturbation signature
\begin{equation}
\Delta x_{i,d,c,g}=\log(1+x_{i,d,c,g})-|C_c|^{-1}\sum_{j\in C_c}\log(1+x_{j,\mathrm{DMSO},c,g}),
\label{eq:signature}
\end{equation}
where $C_c$ contains DMSO cells for cell line $c$. Values are clipped to $[-5,5]$; the 256 genes with largest $|\Delta x_g|$ are retained from a 4,184-gene input universe. The model ranks (1) genes in a 278-entry target vocabulary and (2) recorded entries in a fixed 379-compound molecular bank.

We retain non-control Tahoe-100M drug--cell-line groups with at least 1,000 cells and a mapped target annotation. The 11,673 pre-sampling groups are split 90/10 within each compound using seed 42, yielding 10,505 training and 1,168 validation groups. No complete drug--cell-line group is shared across partitions, although compound and cell-line identities may recur individually. The protocol thus tests recognition of a known compound from cells in a held-out drug--cell-line context rather than compound-cold generalization. Of the 379 bank entries, 264 have mapped annotations; two of these have zero molecular vectors and cannot generate queries. Thus no more than 262 identities generate queries, leaving 117 distractor-only entries under this upper bound, while all 379 entries are ranked. Tahoe target lists were proposed by GPT-4o from compound names and checked against MedChemExpress by the dataset authors \citep{ref15}; they are metadata, not exhaustive causal ground truth. The public dataset is released under CC0-1.0.

\begin{figure}[ht]
\centering
\input{figures/pipeline}
\caption{GeneSpeak-FP maps a single observed perturbation signature to target and compound retrieval spaces. The closed output banks make ranking auditable, but also bound the result to known library entries and metadata annotations.}
\label{fig:pipeline}
\end{figure}

\paragraph{Model.}
The sequence [CLS], [ORGAN], $g_1,\ldots,g_L$ combines each gene-token embedding with a linear projection of its signed $\Delta x_g$. The model (5,353,985 trainable parameters) uses a four-layer, eight-head Transformer (width 256; feed-forward width 1,024; dropout 0.1) with no positional embeddings and therefore acts as an attention-based set encoder \citep{ref5}. The final CLS state feeds a 256-dimensional target head and a $256\to1{,}024\to768$ molecular head.

Target scores are cosine similarities against trainable embeddings for the 278 target genes. Training combines cosine alignment to the mean positive-target embedding, multi-label BCE, and multi-positive ranking. Molecular training uses symmetric in-batch contrastive learning against fixed 768-dimensional molecular vectors plus paired cosine alignment. The total objective is
\begin{equation}
\mathcal L=\mathcal L_{\mathrm{target}}+0.05\bigl(\mathcal L_{\mathrm{contrastive}}+0.5\mathcal L_{\mathrm{align}}\bigr).
\label{eq:joint}
\end{equation}
All non-annotated genes in the closed target vocabulary are optimized as negatives, although non-annotation is not biological evidence of non-target status. Full loss definitions, masking, and negative-set details are in Appendix~\ref{app:model}.

\section{Evaluation}
\paragraph{Protocol.}
Training uses AdamW, a cosine OneCycle schedule, mixed precision, gradient clipping, and stochastic condition-weighted sampling. The evaluated checkpoint is the final state after 16 full nominal epochs and 2,032 microbatches of epoch 17; it was not selected by validation performance. A nominal epoch consists of 7,000 generated microbatches rather than an exhaustive data pass. Validation samples 300 microbatches of 128 cells (38,400 queries) from the fixed condition pool. We report target Recall/Precision@$K$ and compound Hit@$K$, MRR, and rank. Here, $K$ is the number of highest-ranked candidates retained. Target Recall@$K$ is the fraction of a query's annotated targets retrieved in those $K$ positions, and Precision@$K$ is the fraction of those positions occupied by annotated targets; both are averaged across eligible queries. Compound Hit@$K$ is the fraction of queries whose recorded compound appears in the top $K$. MRR is the mean reciprocal rank of that compound. Appendix~\ref{app:protocol} gives the complete configuration.

The comparison includes a Gaussian query-vector control and three bag-of-genes (BoG) controls: prototype matching, a learned linear molecular head, and nearest-neighbor retrieval. These controls reuse learned embeddings or other parts of the GeneSpeak-FP representation and are diagnostic rather than independently trained baselines. They test whether the reported signal is explained by random geometry or simple post-hoc aggregation; they do not show that self-attention, multi-task training, or the proposed architecture is necessary.

\begin{table}[ht]
\caption{Validation comparison. Target metrics rank 278 annotated genes; compound metrics rank all 379 recorded bank entries. The trained model shows clear closed-library signal, but the controls are diagnostic rather than a complete external benchmark.}
\label{tab:comparison}
\centering
\begin{tabular}{lrrrrrr}
\toprule
Method & R@5 & R@10 & P@10 & H@1 & H@5 & H@10\\
\midrule
Uniform rank (analytic) & .0180 & .0360 & -- & .0026 & .0132 & .0264\\
Gaussian vector & .0145 & .0312 & .0069 & .0022 & .0111 & .0223\\
BoG prototype & .0073 & .0167 & .0051 & .0022 & .0194 & .0366\\
BoG linear head & .0072 & .0166 & .0050 & .0029 & .0148 & .0325\\
kNN-BoG & .0072 & .0162 & .0051 & .0035 & .0170 & .0340\\
GeneSpeak-FP & .2930 & .4090 & .0648 & .1288 & .2710 & .3420\\
\bottomrule
\end{tabular}
\par\smallskip
\begin{minipage}{\linewidth}\small
R = target recall; P = target precision; H = compound hit rate. GeneSpeak-FP target Recall@20 is 0.5444 in the \emph{separate extended pass} (Table~\ref{tab:extended}); this value is not from the comparison pass above.
\end{minipage}
\end{table}

\paragraph{Results.}
Across the 38,400-query comparison pass (Table~\ref{tab:comparison}), GeneSpeak-FP places the correct recorded compound in the top 10 for 34.2\% of cells and at rank one for 12.9\%; it recovers 40.9\% of annotated targets within the top 10. Relative to analytic chance, compound Hit@10 and target Recall@10 improve by approximately $13\times$ and $11\times$, respectively, while compound Hit@1 is nearly $49\times$ higher. Compound Hit@10 is also $9.3\times$ the strongest evaluated BoG control. A separately executed extended Monte Carlo pass yields nearly identical aggregate metrics: target Recall@10/20 is 0.4075/0.5444, compound Hit@1/10 is 0.1286/0.3429, and MRR is 0.2048. Small differences across passes are expected because individual validation cells are drawn stochastically. The reported model was trained with a fixed seed of 42, but no uncertainty interval across independently retrained seeds is available.

The analytic uniform-rank references are $K/278$ for target Recall@$K$ and $K/379$ for compound Hit@$K$: these are the expected scores if all candidates are ranked uniformly at random. A uniform-ranking reference restricted to the maximum 262 query-eligible identities would instead have Hit@10 of $10/262=0.0382$; the reported evaluation ranks all 379 entries. Gaussian-vector scores need not match those values because the fixed banks are geometrically nonuniform and query identities are a nonuniform subset. An audit of the $379\times768$ molecular tensor finds two duplicate-vector pairs that coincide with duplicate canonical SMILES and two zero rows. Duplicate vectors can create order-dependent ties for exact recorded identity; zero-vector entries remain distractors but are not queries. Appendix~\ref{app:extended} reports the extended metrics, and Appendix~\ref{app:audit} reports the audit.

\section{From retrieval to drug-discovery evidence}
\paragraph{What the experiment supports.}
Across 38,400 cell-level queries, GeneSpeak-FP recovers 34.2\% of recorded compounds within the top 10 of a 379-entry bank and 40.9\% of annotated targets within the top 10 of a 278-gene vocabulary. These values exceed analytic chance by approximately $13\times$ and $11\times$, and compound Hit@10 is more than $9\times$ the strongest evaluated post-hoc control. The closely matching extended pass provides a pass-to-pass sampling-consistency check for the retained checkpoint, although it is not an independent replication or a training-seed robustness result. Together, these gaps establish strong closed-library identification signal for held-out drug--cell-line contexts and support retrospective screen organization and candidate-space narrowing. The current controls do not, however, establish an architecture-level advantage over independently trained linear, MLP, or set-pooling models.

\paragraph{What it does not support.}
Because compounds recur across partitions and target lists are constant at the compound level, the target head may recognize a familiar compound and recover memorized annotations. This possibility limits interpretation as direct biological target inference, but it does not erase the compound-conditioned metadata-retrieval result: such a route still requires identifying a familiar compound record from a single cell in a held-out drug--cell-line group. The target metrics therefore support retrieval of compound-associated annotations, not causal targets. The evaluation does not test unseen compounds, unseen cell lines, drug combinations, causal mechanisms, efficacy, toxicity, or clinical utility. Dose, time, plate, and replicate identifiers are absent from the grouping and model; DMSO references are cell-line-wide rather than plate-matched. These omissions matter because technical or exposure-related signatures could support retrospective identity ranking without transporting to a new campaign.

\paragraph{Evidence needed for translation.}
A controlled next evaluation should freeze train/validation/test and cell-level manifests; compare independently trained linear, MLP, and mean-pooling models; include compound-ranking-to-target and single-task baselines; repeat training across seeds with condition- or compound-clustered uncertainty; and evaluate compound-cold and cell-line-cold splits. Plate-matched controls and explicit dose/replicate handling are required to probe confounding. Because per-cell class overlap can obscure population-level identifiability \citep{ref9}, retrieval should also be reported as a function of the number of cells aggregated per held-out condition. Finally, a blinded prospective assay on new experimental samples is needed before using rankings to allocate wet-lab resources. Failure at any stage would itself identify where the retrospective-to-prospective translation breaks.

\section{Conclusion}
GeneSpeak-FP establishes a strong closed-library retrieval result from individual single-cell perturbation signatures: compound Hit@10 reaches 0.343 over 379 records, while target Recall@10 reaches 0.408 over 278 metadata genes, with a closely matching separate Monte Carlo pass. The result provides an auditable reference point for retrospective candidate-space narrowing and motivates the next stage of evaluation across independent baselines, cold splits, population aggregation, technical covariates, and prospective samples.

%% file: figures/pipeline.tex
% Editable, native-width pipeline: 394 pt x 150 pt, including stroke margins.
% Required TikZ libraries: arrows.meta (positioning is not required).
\begin{tikzpicture}[
  x=1pt,y=1pt,
  font=\sffamily\fontsize{7.5}{9}\selectfont,
  every node/.style={inner sep=0pt,outer sep=0pt},
  flow/.style={-{Stealth[length=3.6pt,width=3.3pt]},line width=0.65pt},
  panel/.style={rounded corners=4pt,line width=0.6pt},
  title/.style={font=\sffamily\bfseries\fontsize{9}{10.5}\selectfont},
  label/.style={font=\sffamily\bfseries\fontsize{7}{8.5}\selectfont}
]
  \definecolor{gsnavy}{HTML}{16476B}
  \definecolor{gsteal}{HTML}{008B9B}
  \definecolor{gsamber}{HTML}{A65E00}
  \definecolor{gstext}{HTML}{1D2E3D}

  % The explicit bounds prevent downstream changes in font metrics from
  % making this fragment wider than the manuscript's 396 pt text block.
  \path[use as bounding box] (-0.5,-0.5) rectangle (393.5,149.5);

  \node[title,anchor=west,text=gsnavy] at (0,143)
    {GeneSpeak-FP: one perturbation, two retrieval spaces};

  % Input and shared set encoder.
  \draw[panel,draw=gsnavy,fill=gsnavy!5] (0,40) rectangle (95,128);
  \node[label,text=gsnavy,align=center,text width=87pt] at (47.5,117)
    {CELL-LEVEL QUERY};
  \node[text=gstext,align=center,text width=87pt,anchor=north] at (47.5,104)
    {log1p(treated cell)\\
     $-$ cell-line DMSO mean\\
     (after log1p)\\[3pt]
     Clip to $[-5,5]$\\
     Keep 256 largest\\
     absolute changes};

  \draw[panel,draw=gsteal,fill=gsteal!5] (113,40) rectangle (241,128);
  \node[label,text=gsteal,align=center,text width=118pt] at (177,117)
    {GENE-TOKEN SET ENCODER};
  \node[text=gstext,align=center,text width=118pt,anchor=north] at (177,101)
    {[CLS], [ORGAN], gene tokens\\[3pt]
     4 layers; 8 attention heads\\
     $d=256$; FF $=1{,}024$\\[3pt]
     No positional embeddings};

  % Distinct retrieval spaces and their fixed candidate banks.
  \draw[panel,draw=gsamber,fill=gsamber!6] (263,88) rectangle (393,128);
  \node[label,text=gsamber,align=center,text width=120pt] at (328,119)
    {ANNOTATED TARGETS};
  \node[text=gstext,align=center,text width=120pt,anchor=north] at (328,110)
    {256-D head ranks 278 Tahoe\\
     target-metadata genes};

  \draw[panel,draw=gsnavy,fill=gsnavy!5] (263,40) rectangle (393,80);
  \node[label,text=gsnavy,align=center,text width=120pt] at (328,71)
    {RECORDED COMPOUNDS};
  \node[text=gstext,align=center,text width=120pt,anchor=north] at (328,62)
    {$256\!\rightarrow\!1{,}024\!\rightarrow\!768$ head\\
     Ranks 379 fixed molecular vectors};

  \draw[flow,draw=gsnavy] (95,84) -- (113,84);
  \draw[flow,draw=gsamber] (241,84) -- (253,108) -- (263,108);
  \draw[flow,draw=gsnavy] (241,84) -- (253,60) -- (263,60);

  % One shared training objective; the second line continues the first.
  \draw[panel,draw=gsnavy!25,fill=gsnavy!3] (0,0) rectangle (393,28);
  \node[text=gstext,anchor=west] at (9,19)
    {\textbf{Joint objective:} target cosine $+\,0.05$ BCE $+$ multi-positive ranking};
  \node[text=gstext,anchor=west] at (9,8.5)
    {$+\,0.05\times$ (symmetric contrastive $+\,0.5$ cosine alignment)};
\end{tikzpicture}

%% file: references.tex
% Bibliography transcribed from the current OpenReview PDF, pages 5--6.
% Requires the url package (or hyperref); xurl is recommended for URL wrapping.
% Numeric keys preserve the current PDF's citation order exactly.

%% file: appendices.tex
\section{Dataset, labels, and split}\label{app:data}
\paragraph{Scope.}
Table~\ref{tab:scope} summarizes the retained experimental scope. Target labels come from the \texttt{targets} field of the Tahoe drug metadata and are mapped to Tahoe genes by Ensembl identifier or case-insensitive gene symbol. Among the 264 compounds with at least one mapped target, the number of labels ranges from 1 to 13 (mean 2.08, median 1, interquartile range 1--3). There are 142 compounds with one mapped target and 122 with multiple targets. For 1, 2, 3, 4, 5, 6, 7, 8, 9, 10, 12, and 13 labels, the respective compound counts are 142, 53, 40, 10, 6, 4, 3, 1, 1, 1, 1, and 2.

\begin{table}[ht]
\caption{Experimental scope of the evaluated dataset and banks. Counts of condition pairs describe the pre-sampling split.}
\label{tab:scope}
\centering
\begin{tabular}{lr}
\toprule
Item & Count\\
\midrule
Compound bank $|D|$ & 379\\
Compounds with mapped targets & 264\\
Maximum query-eligible compounds & 262\\
Closed target vocabulary $|T|$ & 278\\
Input gene universe & 4,184\\
Filtered drug--cell-line groups & 11,673\\
Training groups & 10,505\\
Validation groups & 1,168\\
\bottomrule
\end{tabular}
\end{table}

Queries are sampled only from groups whose compounds have at least one mapped target and a nonzero molecular vector. Other library entries remain distractors, including 115 compounds without mapped targets and two mapped-target entries with zero molecular vectors. The exact number of query-positive identities and the exact sampleable group counts cannot be recovered from the retained aggregate outputs, so Table~\ref{tab:scope} reports an upper bound of 262 identities and pre-sampling group counts.

\paragraph{Condition definition and controls.}
In this paper, a ``condition group'' means a drug--cell-line grouping, not the dataset's full drug--concentration treatment condition. Concentration, sample, plate, and replicate identifiers are not part of the grouping key or model input. DMSO cells are log-transformed and pooled within each cell line; a global DMSO mean is used only as a fallback. The retained aggregate outputs do not record how often the fallback was used. Plate-matched controls are not implemented.

The 4,184-gene input universe is the union of 4,000 genes selected by variance among DMSO cells and all mapped target genes. The retained preprocessing artifact does not permit a complete audit of whether the variance selection was restricted to training data. Each query keeps the 256 genes with largest absolute change and sorts them by token identifier to produce deterministic ordering.

\paragraph{Dataset access and license.}
Tahoe-100M is publicly available from the dataset authors on Hugging Face \citep{ref15} under the CC0-1.0 license. The experiments use public cancer cell-line perturbation data and do not involve patient-level clinical records or human-subject intervention data.

\section{Model and objective details}\label{app:model}
Sequences are right-padded to 258 positions (CLS, ORGAN, and up to 256 genes) with \texttt{PAD\_ID=0} and zero scalar values. The binary attention mask equals one at the two special-token positions and at populated gene positions, and zero at padding positions. The PyTorch Transformer receives \texttt{src\_key\_padding\_mask=(attention\_mask==0)}. The token embedding uses \texttt{padding\_idx=0}. A coarse organ embedding derived from cell-line metadata is added at the ORGAN position.

\paragraph{Target loss.}
Let $z_T$ be the target-head output and $e_g$ the trainable embedding of target gene $g$. Scores use cosine similarity. For a compound's positive target set $P$, let $\bar e_T$ be the normalized mean positive embedding. The target loss is
\begin{equation}
\mathcal L_T=1-\cos(z_T,\bar e_T)+0.05\mathcal L_{\mathrm{BCE}}+\mathcal L_{\mathrm{rank}}.
\label{eq:target}
\end{equation}
The BCE term uses all positive labels and cosine logits with temperature 0.15. Although the configured caps are 2,048 BCE negatives and 1,024 ranking negatives, only 265--277 non-positive labels exist per compound in the 278-entry vocabulary. The executed run therefore uses every non-positive target in both terms. The ranking term retains at most four positives and is
\begin{equation}
\mathcal L_{\mathrm{rank}}=-\frac{1}{|P|}\sum_{p\in P}\log\frac{\exp(s_p/\tau_r)}{\sum_{q\in P\cup N}\exp(s_q/\tau_r)},\qquad \tau_r=0.15,
\label{eq:ranking}
\end{equation}
where $N$ is the complete set of non-positive labels.

\paragraph{Molecular loss.}
For a microbatch, let $Z$ be normalized predicted molecular vectors and $V$ the normalized fixed molecular vectors. The symmetric contrastive term is
\begin{equation}
\mathcal L_{\mathrm{contrastive}}=\frac12\bigl[\mathrm{CE}(ZV^\top/\tau_M,I)+\mathrm{CE}(VZ^\top/\tau_M,I)\bigr],
\label{eq:contrastive}
\end{equation}
where $\tau_M$ is learned from an initial value of 0.10. The paired alignment term is $\mathcal L_{\mathrm{align}}=1-B^{-1}\sum_i\cos(z_{M,i},v_i)$. Contrastive candidates remain within each 128-record microbatch rather than spanning the four microbatches combined by gradient accumulation.

\section{Training and evaluation protocol}\label{app:protocol}
Optimization uses AdamW with weight decay 0.01, $\epsilon=10^{-8}$, and $\beta_2=0.999$. A cosine OneCycle schedule has 35,000 scheduled update steps, maximum learning rate $10^{-4}$, \texttt{pct\_start=0.05}, \texttt{div\_factor=10}, and \texttt{final\_div\_factor=100}. The initial and minimum learning rates are $10^{-5}$ and $10^{-7}$. With OneCycle momentum enabled, $\beta_1$ cycles from 0.95 to 0.85 and back to 0.95. Training uses mixed precision, gradient clipping at 1.0, microbatches of 128, and four-step gradient accumulation.

A nominal epoch contains 7,000 stochastically generated microbatches (896,000 sampled records), giving 1,750 scheduled optimizer-update slots. The sampler is designed to use distinct recorded compound identities in each microbatch. Training and validation use inverse-square-root group weights to reduce the influence of groups with more cells. Each record is generated by selecting a group and then drawing one treated cell from an indexed Parquet row group.

The group split and the main Python, NumPy, and PyTorch random-number generators use seed 42; worker-specific sampling uses fixed local seeds. The checkpoint retains Python, NumPy, PyTorch CPU, and CUDA random states. Some DataLoader sampling depends on runtime-managed state, TF32 was enabled, and deterministic CUDA algorithms were not enforced. The protocol is therefore not a bitwise replay specification, and no independently retrained seeds are available.

Notebook progress logs sum to 47:24:18 (47.405 hours) of training-loop time across the 16 completed epochs and the partial seventeenth epoch. This excludes validation, preprocessing, indexing, data loading outside the logged loops, and setup. The exact accelerator model and memory, end-to-end wall-clock time, and total exploratory compute were not retained.

\paragraph{Checkpoint.}
The evaluated checkpoint was saved after interruption during epoch 17, when the progress display showed 2,032 of 7,000 microbatches, following 16 complete nominal epochs. Internal BatchNorm and scheduler counters are consistent with this point in training. The checkpoint stores model weights, AdamW state, OneCycle scheduler state, AMP scaler state, configuration and lookup mappings, aggregate validation metrics, and Python, NumPy, PyTorch CPU, and CUDA random states. It therefore supports substantially more faithful stateful continuation than weights alone, although missing data artifacts still prevent exact end-to-end replay. The checkpoint was not selected by validation performance or early stopping; its validation dictionary belongs to the last completed validation pass at epoch 16. Target ranking was active from the first epoch.

\begin{figure}[ht]
\centering
% Original vector panels cropped from page 9 of the authoritative current PDF.
% Stacked and enlarged for readable axis/legend text; curves were not re-estimated.
\begin{minipage}{0.82\linewidth}
\sffamily\small\bfseries GeneSpeak-FP: retained one-run validation trajectory\\
\normalfont\footnotesize 16 completed epochs | stochastic validation passes
\end{minipage}
\par\smallskip
\includegraphics[width=0.82\linewidth]{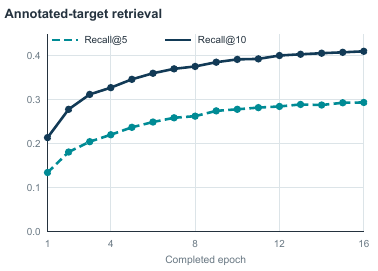}
\par\smallskip
\includegraphics[width=0.82\linewidth]{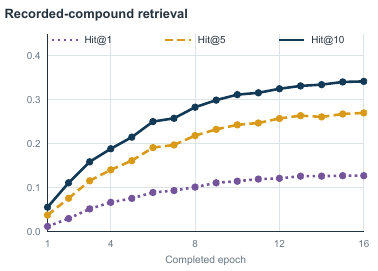}
\caption{Retained one-run validation trajectory across the 16 completed epochs. Retrieval metrics improve broadly over training, while small local reversals can occur because each epoch uses a stochastic validation pass. The figure does not represent repeated-seed uncertainty.}
\label{fig:trajectory}
\end{figure}

Across the 16 completed epochs, the logged mean training objective decreased monotonically from 6.4425 to 5.0031. Validation target Recall@10 increased from 0.2140 to 0.4105, while compound Hit@10 increased from 0.0555 to 0.3420. Because validation draws are stochastic and this is one training seed, the trajectory in Figure~\ref{fig:trajectory} is descriptive rather than an uncertainty estimate.

\paragraph{Metrics.}
Target rankings contain only the 278-gene vocabulary. Recall@$K$ and Precision@$K$ are averaged across samples with at least one positive target. Compound retrieval ranks all 379 molecular vectors and reports Hit@$K$, mean reciprocal rank (MRR), and rank summaries. The analytic uniform-ranking expectations are target Recall@$K=K/278$ and compound Hit@$K=K/379$.

\paragraph{Diagnostic controls.}
The Gaussian-vector control scores randomly drawn queries against the fixed banks; it is not a uniform random permutation because bank geometry is nonuniform. The BoG prototype, linear molecular, and nearest-neighbor controls reuse learned embeddings or other components of the GeneSpeak-FP representation. For the linear molecular control, compound retrieval uses a learned projection of the delta-weighted BoG vector. Target rankings are computed post hoc by cosine similarity between its normalized pre-projection representation and the fixed target-gene embedding bank. Token and organ embeddings are initialized from GeneSpeak-FP and optimized only through molecular-embedding MSE; there is no target-specific head or loss. These are diagnostic controls, not independent competitors trained from raw inputs.

\section{Extended evaluation}\label{app:extended}
\begin{table}[ht]
\caption{Exact values from the 38,400-query extended Monte Carlo validation output. The batch-median mean is an average of batch-level medians and is not a global median over all queries.}
\label{tab:extended}
\centering
\begin{tabular}{lrlr}
\toprule
Target metric & Value & Compound metric & Value\\
\midrule
Recall@5 & 0.2925 & Hit@1 & 0.1286\\
Recall@10 & 0.4075 & Hit@5 & 0.2682\\
Recall@20 & 0.5444 & Hit@10 & 0.3429\\
Precision@10 & 0.0644 & MRR & 0.2048\\
mAP@10 & 0.2101 & Mean rank & 72.7407\\
NDCG@10 & 0.2647 & Batch-median mean & 32.1267\\
\bottomrule
\end{tabular}
\end{table}
The comparison table in the main paper and this extended evaluation are separate stochastic passes. Because a fixed cell-level manifest and per-query outputs were not retained, they cannot be paired query by query and small differences should not be interpreted as model changes.

\section{Initialization and candidate-bank audit}\label{app:audit}
\paragraph{Gene initialization.}
The external gene-embedding warm start was stored without special-token rows, whereas the retrieval implementation indexed it with Tahoe token identifiers. This introduced a fixed row offset at initialization. The table remained fully trainable; the study therefore treats it only as a generic warm start and does not attribute performance to correctly aligned biological semantics. A future rerun should correct the mapping and compare corrected, random, and no-warm-start initializations under shared seeds and manifests.

\paragraph{Molecular bank.}
The retained molecular tensor has shape $379\times768$ and contains no nonfinite values. In metadata-row order, its two nonzero duplicate-vector pairs coincide with duplicate canonical SMILES: crizotinib/(S)-Crizotinib and Doxorubicin (hydrochloride)/Epirubicin (hydrochloride). Its two zero rows coincide with entries lacking canonical SMILES, Sacubitril/Valsartan and Verteporfin. The executed pipeline attached vectors to drug names and built the evaluation bank in lexicographically sorted drug-name order.

Distinct recorded IDs with identical vectors are treated as separate one-hot pairs by the historical contrastive objective. If both occur in a microbatch, they form false-negative pairs even though their molecular vectors are identical; the frequency was not logged. At evaluation, duplicate vectors produce tied scores. The retained outputs do not document a tie-aware multi-positive policy, so exact-record ranking for these pairs can depend on candidate order and sorting behavior. Structure-level multi-positive scoring would be more appropriate for this subset.

The molecular tensor was cross-checked against the 379-row drug metadata table through its size, duplicate patterns, zero rows, canonical SMILES, and executed ordering. Because the original named molecular-bank object and exact evaluation manifest were not retained, this is an audit cross-check rather than an independently archived row-to-record manifest.

\clearpage
\section{Artifacts, impact, and responsible use}\label{app:artifacts}
Retained project artifacts include the trained retrieval checkpoint, fixed molecular embedding tensor, gene and cell embedding arrays, metadata tables, executed notebook code, model configuration, and aggregate evaluation outputs. The checkpoint contains model, optimizer, scheduler, and AMP-scaler states, lookup mappings, aggregate metrics, and Python, NumPy, PyTorch CPU, and CUDA random states. These materials support inspection of the architecture, candidate banks, training state, and aggregate results. The exact sampled-cell manifest, per-query predictions, original DMSO preprocessing artifact, and the checkpoint and pooling configuration used to generate the molecular embeddings were not retained. Consequently, the downstream model and training state can be inspected, but the original preprocessing and stochastic draws cannot be reproduced exactly. No public project-artifact link is provided in this manuscript.

The dataset authors and dataset card are credited, and the public Tahoe dataset license is CC0-1.0. The precise provenance and license of the unavailable historical gene-initialization and molecular-encoder checkpoints cannot be fully verified from the retained files; this should be resolved before any public release of derived artifacts.

The intended use is retrospective dataset exploration and narrowing of a predefined candidate bank. A positive impact is more efficient organization of existing in vitro perturbation screens. The principal risk is treating a ranking as causal, clinical, or therapeutic evidence. Appropriate safeguards are interpretive and procedural: disclose the closed-library scope, label provenance, and known failure modes; require independent biological and prospective validation; and prohibit use for clinical decision-making or as evidence that a compound is safe, effective, or mechanistically causal.